\documentclass[runningheads]{llncs}
\usepackage{amsmath, amssymb}
\usepackage[T1]{fontenc}
\usepackage{graphicx}
\usepackage{hyperref}
\usepackage{color}

\begin{document}
\title{Depth-Wise Representation Development Under Blockwise Self-Supervised Learning for Video Vision Transformers}
\titlerunning{Depth-Wise Representation Development Under BWSSL for Video ViTs}
%
\author{Jonas Römer\inst{1}\orcidID{0009-0008-3236-7647} \and
Timo Dickscheid\inst{2,3,4}\orcidID{0000-0002-9051-3701}}
\authorrunning{J. Römer \and T. Dickscheid}
%
\institute{Heinrich Heine University, Düsseldorf, Germany \and
Institute for Computational Visualistics, University of Koblenz, Germany \and 
Institute of Neuroscience and Medicine (INM-1), Research Centre Jülich, Germany \and
Helmholtz AI, Research Centre Jülich, Germany\\
\email{Jonas.Roemer@hhu.de}\\\email{dickscheid@uni-koblenz.de}}
\maketitle              

\begin{abstract}
End-to-end backpropagation couples all layers through a global error signal, enabling coordinated learning but requiring long-range credit assignment. Motivated by recent progress in blockwise self-supervised learning (BWSSL), we ask whether masked video transformers can be trained without end-to-end backpropagation. Applying BWSSL to masked video modeling remains relatively underexplored and must handle spatiotemporal context and long-range temporal structure. More broadly, analyses that compare BWSSL and end-to-end training in terms of learning dynamics and depth-wise representation development remain sparse.
We apply blockwise learning to a masked autoencoding video vision transformer by partitioning the encoder into blocks, each of which is optimized with a local masked reconstruction loss. Across model sizes and partition granularities, training converges and yields representations close to matched end-to-end baselines under linear-probe and retrieval proxies. In order to compare intermediate representations, we analyze depth-wise decodability, inter-block similarity, and patch-level diagnostics. Blockwise training exposes higher-level structure earlier, while later blocks saturate and operate in a more geometry-preserving regime. It can also induce token-level shifts consistent with stronger early mixing that pooled metrics can miss. These findings point to late-block saturation and interface formation as contributors to the remaining gap.

Evaluation Code: \url{https://github.com/JosRor/BWSSL-for-Video-ViTs}

\end{abstract}

\section{Introduction}
End-to-end (E2E) backpropagation is the standard method for training deep neural networks, coupling all layers through a single global error signal and requiring long-range credit assignment. Motivated by biological constraints and the view that intelligence may rely on strong inductive structure together with more local learning mechanisms, there is growing interest in local-learning alternatives \cite{zadorCatalyzingNextgenerationArtificial2023,lillicrapBackpropagationBrain2020a}.

A pragmatic compromise is blockwise learning (BWL) with gradient isolation: the network is partitioned into blocks, gradients are stopped at block boundaries, and each block is optimized with its own objective \cite{lowePuttingEndEndtoEnd2019,siddiqui2024blockwise,denoodtSmoothInfoMaxEasier2026,karkarModulewiseTrainingNeural2023a,huangDeInfoRegDecoupledLearning2025}. When paired with strong self-supervised objectives, blockwise self-supervised learning (BWSSL) scales to ImageNet pretraining and can approach matched E2E performance, with a small residual gap under strict isolation \cite{siddiqui2024blockwise,xiongLoCoLocalContrastive2020}.

However, most large-scale evidence is for static images and often uses CNN backbones \cite{siddiqui2024blockwise,xiongLoCoLocalContrastive2020}. The behavior of BWSSL in transformer backbones, with masked reconstruction objectives and video data is less well studied. Videorepresentation learning introduces spatiotemporal structure and motion. Two questions remain underexplored: (i) can gradient-isolated BWSSL train video vision transformers with a masked reconstruction objective to near-E2E representation quality, and (ii) how do intermediate representations differ between BWSSL and E2E across network depth? 

We address these questions by applying BWSSL to a Vision Transformer (ViT) \cite{dosovitskiyImageWorth16x162021}. We split the encoder into $K$ blocks, optimize each block with a local VideoMAE masked-reconstruction loss \cite{tongVideoMAEMaskedAutoencoders2022a}, and compare against matched E2E baselines across model sizes and partition granularities. To characterize differences beyond the final representation, we perform depth-resolved analyses of (i) linear decodability, (ii) representational similarity across depth (CKA) \cite{kornblithSimilarityNeuralNetwork2019a}, and (iii) patch/token-level diagnostics, localizing where information becomes accessible and how representations evolve under BWSSL.
The main contributions of this study are:
\begin{enumerate}
    \item We build a BWSSL pipeline for VideoMAE-style video ViTs and benchmark it against matched E2E training across model sizes and block granularities.
    \item We provide evidence that BWSSL converges reliably and approaches E2E representation quality for downstream tasks, with a small residual gap.
    \item We utilise a depth-resolved diagnostic suite (decodability, CKA, and token analyses) to compare intermediate representations under BWSSL vs. E2E.
    \item We uncover consistent depth-wise differences, namely earlier linearly accessible structure and late-block saturation with geometry-preserving updates.
\end{enumerate}

\section{Related Work}
We situate our work within blockwise learning, local objectives for stable training, and depth-wise representation analyses in blockwise systems and ViTs.

\paragraph{Gradient-isolated blockwise learning and BWSSL.}
Gradient-isolated blockwise learning is a module-wise generalization of greedy/layer-wise training, where network blocks are optimized with local objectives and gradients are stopped across block boundaries \cite{bengioGreedyLayerWiseTraining2007,belilovskyGreedyLayerwiseLearning2019,lowePuttingEndEndtoEnd2019}. With strong self-supervised objectives, such blockwise training can approach matched end-to-end (E2E) performance. LoCo narrows the gap by overlapping local modules, which effectively deepens the local decoder and provides implicit longer-range learning signals \cite{xiongLoCoLocalContrastive2020}. Siddiqui et al. demonstrate ImageNet-scale BWSSL in CNNs by training ResNet-50 blocks with a redundancy-reduction objective (Barlow Twins), achieving linear-probe accuracy within \(\approx  1.1\%\) of an E2E baseline \cite{siddiqui2024blockwise}. Beyond CNNs, recent work has extended gradient-isolated blockwise training to ViTs under masked image modeling via per-block decoders and block-local reconstruction losses \cite{luoBIMBlockWiseSelfSupervised2023}. However, gradient-isolated BWSSL has been studied primarily in static-image settings, and its application to masked video modeling with video transformers and depth-resolved analyses of representation development under such training appears comparatively limited.

\paragraph{Supervised BWL and reconstruction-style auxiliaries.}
In locally supervised blockwise training, naive per-block classifiers can induce early-layer information collapse, where later-relevant information is discarded prematurely \cite{wangInfoProLocallySupervised2025,sakamoto2024endtoend}. InfoPro mitigates this by augmenting local supervision with an information-preservation term implemented via a reconstruction-style auxiliary objective \cite{wangInfoProLocallySupervised2025}. More broadly, reconstructing representations rather than raw inputs has been explored as a practical local target, for transformer backbones including latent/feature reconstruction \cite{pathakLocalLearningTransformers2022}. Together, these results motivate reconstructive objectives as a natural fit for stable blockwise training.

\paragraph{Depth-wise representation analysis in blockwise models and ViTs.}
Several works on supervised blockwise learning probe internal behavior beyond final accuracy \cite{suMomentumAuxiliaryNetwork2025a,ma2024scaling,wangInfoProLocallySupervised2025}. 
Sakamoto \& Sato, for example, compare E2E and layer-wise supervised training and report that E2E exhibits \emph{layer-role differentiation} in HSIC-based information dynamics, whereas layer-wise training shows more uniform depth-wise behavior \cite{sakamoto2024endtoend}. 
Separately, depth analyses of E2E ViTs (not BWL) characterize how representations evolve with depth: Raghu et al.\ report relatively high cross-layer representational similarity and smooth depth-wise evolution in ViTs \cite{raghuVisionTransformersSee2021}, and Dorszewski et al.\ show a progression from color/texture concepts to increasingly class-like concepts across layers via neuron labeling \cite{dorszewskiColorsClassesEmergence2026}.
However, there is rather little evidence on how self-supervised blockwise training shapes representations across depth, particularly in transformer-based masked video models. We aim to shed more light on these questions by comparing blockwise and matched E2E VideoMAE training using decodability, similarity, and token-level analyses.

\begin{figure}
\includegraphics[width=\textwidth]{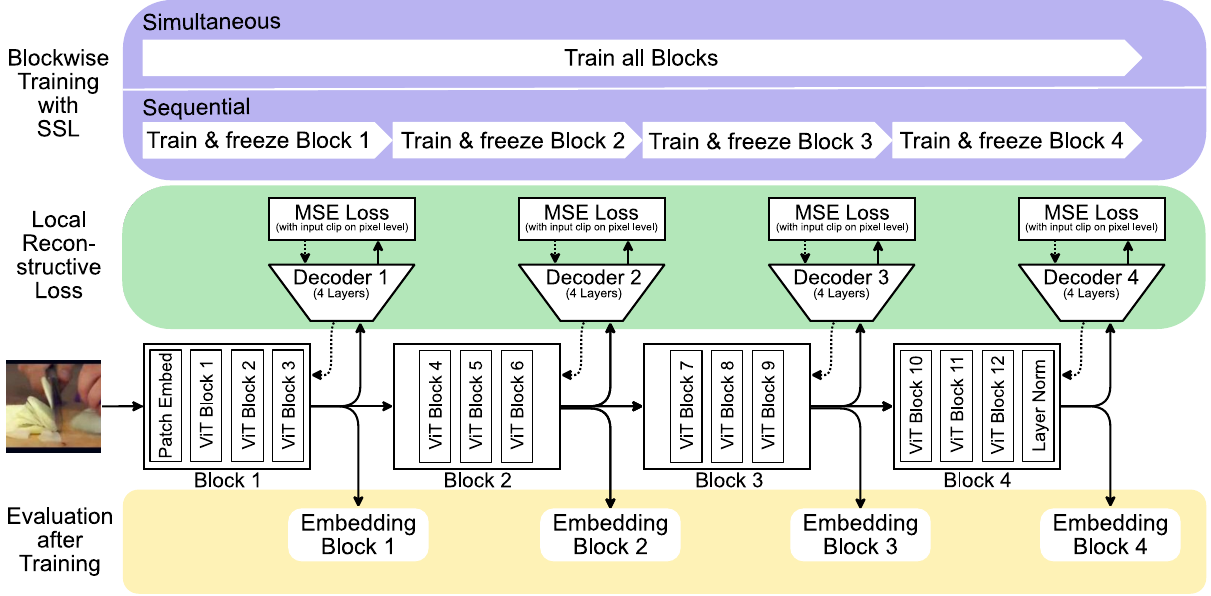}
\caption{
\textbf{Training and Evaluation of the BWSSL model for video data.}
We train a ViT for video data using a masked autoencoding objective. 
The architecture is split into four (this Figure) or six blocks each containing multiple transformer layers. BWSSL training is carried out in sequential \& simultaneous fashion, with MSE reconstruction losses attached to  block outputs. This is compared to a classical E2E training using the last block output. We analyze learned representations by comparing and probing embeddings generated by each block under BWSSL and E2E training.
}
\end{figure}

\section{Blockwise self-supervised training for video data}
\label{sec:BWSSL}

\subsection{Training setup}
We study blockwise self-supervised learning (BWSSL) by splitting a VideoMAE-style video ViT encoder into $K$ gradient-isolated blocks. We write
$f(x)=f_K\circ\cdots\circ f_1(x)$ with $\tilde h_0=x$ for an input clip $x$ and $h_k=f_k(\tilde h_{k-1})$.
To enforce gradient isolation, we insert a stop-gradient operator $\mathrm{sg}(\cdot)$ at block boundaries:
for $k<K$, we pass $\tilde h_k=\mathrm{sg}(h_k)$ to the next block. Thus, losses applied at depth $k$ update only $f_k$ (and its decoder) while leaving forward activations unchanged.

Following VideoMAE, we sample a binary mask on the spatiotemporal token grid and drop masked tokens before the encoder, so each block processes the variable-length sequence of visible tokens. For representation analyses (Sec.~\ref{sec:analysis:downstream} onward), we disable masking and run the encoder on the full token sequence.

Each block $k$ is trained with a local masked-reconstruction loss
\begin{equation}
\mathcal{L}_k = \mathcal{L}_{\text{MAE}}\big(d_k(h_k),\, x;\, m\big),
\end{equation}
where $d_k$ is a VideoMAE-style decoder (Sec.~\ref{sec:architecture}). The mask $m$ is sampled once per clip and reused across all blocks to match the E2E baseline’s masking pattern.

We compare three optimization regimes:
\begin{enumerate}
    \item \textbf{Sequential BWSSL.}
    Blocks are trained in order $k=1,\dots,K$. At stage $k$, we run up to block $k$ and update only $(f_k,d_k)$, with earlier blocks frozen.
    \item \textbf{Simultaneous BWSSL.}
    Each iteration runs a full forward pass with stop-gradients at boundaries, computes $\{\mathcal{L}_k\}_{k=1}^K$, and updates each $(f_k,d_k)$ using its own loss (equivalently, optimizes $\sum_k \mathcal{L}_k$ under gradient isolation).
    \item \textbf{End-to-end (E2E) baseline.}
    We apply the VideoMAE loss only at $h_K$ using $d_K$ and backpropagate through the full encoder.
\end{enumerate}

For each model size, we train E2E and simultaneous BWSSL for $E$ epochs. Sequential BWSSL uses $K$ stages of $E$ epochs: each encoder parameter is updated for $E$ epochs in all regimes, but sequential training makes $K E$ passes over the dataset.
BWSSL differs from E2E in supervision placement in addition to gradient locality due the intermediate reconstruction losses.

\subsection{Training objective}
All regimes use VideoMAE masked video modeling \cite{tongVideoMAEMaskedAutoencoders2022a}. Each decoder predicts tubelet-patch targets, and an MSE reconstruction loss is only computed over masked token positions (averaged over batch, masked tokens, and target dimensions). Masking is applied on the spatiotemporal token grid with tubelets of $(8,16,16)$ and a mask ratio of $90\%$. We use temporal tubelets of 8 mainly for efficiency, reducing encoder and multi-decoder compute under high masking. We do not perform supervised fine-tuning.
Optimization is based on the VideoMAE recipe with AdamW (learning-rate  $1\times10^{-4}$, cosine schedule, weight decay $0.05$).

\subsection{Network architecture}
\label{sec:architecture}
We use a 12-layer ViT with tubelet tokenization \cite{dosovitskiyImageWorth16x162021,touvronTrainingDataefficientImage2021,arnabViViTVideoVision2021}, in DeiT-Tiny and DeiT-Small configurations (embedding dim 192/384, heads 3/6) \cite{touvronTrainingDataefficientImage2021}. We train DeiT-Tiny for $E{=}150$ and DeiT-Small for $E{=}300$ epochs.

For BWSSL, we partition the encoder into either $K{=}4$ blocks (default, matching common practice for comparability) of 3 layers or $K{=}6$ blocks of 2 layers to probe sensitivity to partition granularity. Block 1 includes the tubelet embedding, and block $K$ applies the standard output LayerNorm, while blocks are otherwise identical. (We found no notable differences from inserting LayerNorm at every boundary and keep the standard formulation to match the E2E baseline.)
Each block uses a 4-layer ViT decoder, with a width of 192 and 3 attention heads (fixed across $K$, model sizes, and regimes). E2E uses one decoder, whereas BWSSL uses $K$. Let $P_{\text{enc}}$ and $P_{\text{dec}}$ denote encoder and single-decoder parameters. Then
$P_{\text{E2E}}=P_{\text{enc}}+P_{\text{dec}}$ and $P_{\text{BW}}=P_{\text{enc}}+K P_{\text{dec}}$,
so the relative increase is $(K-1)P_{\text{dec}}/(P_{\text{enc}}+P_{\text{dec}})$.
Thus, larger $K$ can substantially increase parameters and typically also compute, despite identical encoder size. Therefore, even though credit-assignment paths are shorter, memory usage can still increase.

\subsection{Training data}
\label{sec:setup:data}
We train on the official UCF101 train split \cite{soomroUCF101Dataset1012012}. 
We use a frame-based representation as in prior work \cite{hanVideoRepresentationLearning2019b}. Each clip is a contiguous sequence of $T{=}16$ frames with temporal stride $s{=}4$ (span $Ts=64$ original frames). Videos shorter than $Ts$ frames are excluded. We apply random cropping to $128\times128$ and random horizontal flipping. For efficiency, we subsample $30\%$ of the test split ($N\approx1100$) and use center cropping as augmentation.

\section{Comparative Representation Analysis}
We compare BWSSL with matched E2E training by analyzing intermediate and final representations.
Given a trained encoder partitioned into $K$ blocks, we record the block-boundary token representations $h_k$ for $k\in\{1,\dots,K\}$.
When a single vector per clip is required, we form $z_k$ by mean-pooling patch tokens.
We evaluate representations along four axes: (i) downstream performance (linear probing and retrieval), (ii) feature complexity across depth, (iii) representational change across blocks (CKA), and (iv) patch-level detail retention and robustness.

Each model is trained with 3 random initialisations. Metrics that fit an auxiliary model (e.g. linear probes) are also evaluated 3 times per trained network. Unless stated otherwise, metrics operate on $z_k$ and token-level analyses on $h_k$.

\subsection{Downstream task performance metrics}
\label{sec:analysis:downstream}
\begin{figure}
\includegraphics[width=\textwidth]{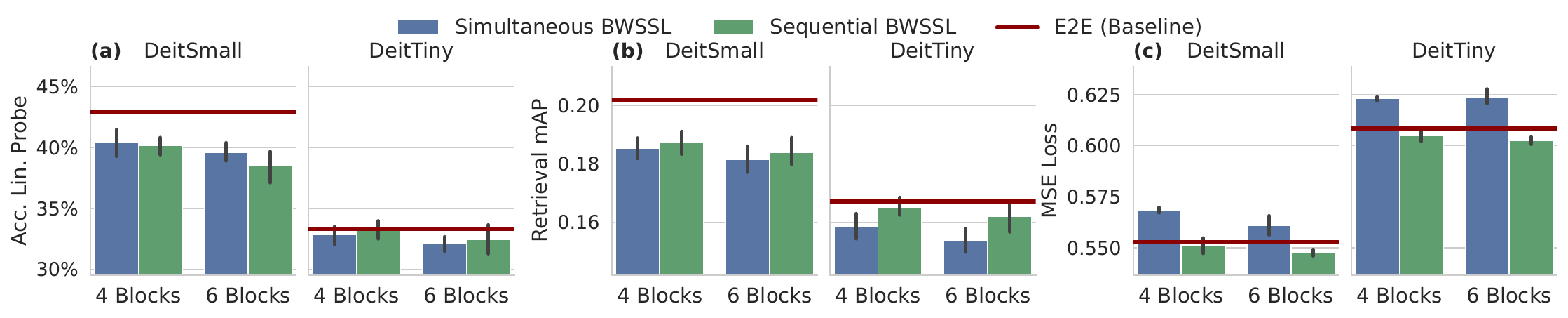}
\caption{\textbf{Downstream task metrics.}
Results for BWSSL ($K{=}4,6$; sequential vs.\ simultaneous) and matched E2E baselines, for DeiT-Tiny and DeiT-Small on UCF101.
(a) Linear-probe accuracy.
(b) Retrieval mAP.
(c) Masked reconstruction MSE.}
\label{fig:key_metrics}
\end{figure}

We evaluate frozen embeddings extracted after pretraining on UCF101.
We report three complementary metrics (\figurename~\ref{fig:key_metrics}): (i) linear-probe accuracy, (ii) kNN retrieval mAP in embedding space, and (iii) masked reconstruction MSE.
We compare training regimes and block granularities from Sec.~\ref{sec:BWSSL}.

\emph{Linear-probe accuracy (multinomial logistic regression).}
Linear probing measures how linearly accessible task information is and serves as a proxy for transfer under limited downstream capacity.
We fit a scikit-learn pipeline (standardization + multinomial logistic regression) on training-set clip embeddings to predict UCF101 labels.
We use $1000$ max iterations and otherwise scikit-learn defaults (L2 penalty, $C{=}1.0$, solver \texttt{lbfgs}).
We report top-1 test accuracy (\figurename~\ref{fig:key_metrics}a).

\emph{kNN retrieval (mAP).}
Retrieval evaluates whether embeddings induce a semantically meaningful neighborhood structure.
For mAP each test clip is a query and the remaining test clips form the gallery. We rank by cosine similarity of pooled embeddings and mark items with the same UCF101 label as relevant (excluding the query).
We compute average precision per query with scikit-learn and report mean AP (mAP) across queries, omitting queries with no other relevant items.
We use mAP because it is cutoff-free and rank-sensitive, whereas top-1 is brittle to near-duplicates (\figurename~\ref{fig:key_metrics}b).

\emph{Reconstruction loss (MSE).}
We report masked reconstruction MSE between decoder predictions and patchified targets, averaged over batch, masked tokens, and patch dimensions.
For BWSSL we report per-block MSE via $d_k$ and for E2E the last block MSE with $d_K$ (no intermediate decoders) (\figurename~\ref{fig:key_metrics}c,\figurename~\ref{fig:cka}a).

\begin{figure}
\includegraphics[width=\textwidth]{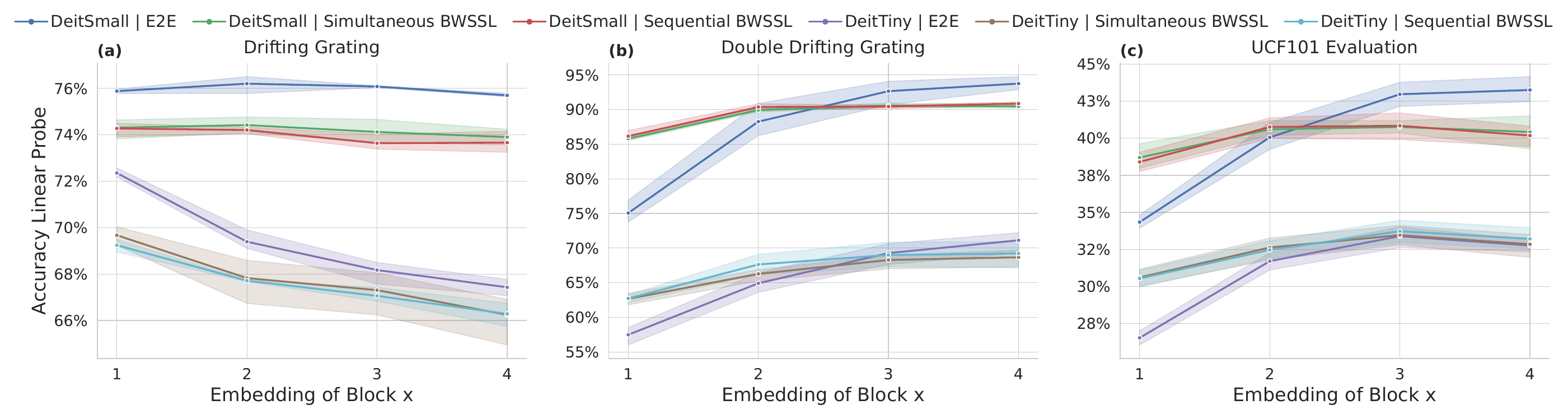}
\includegraphics[width=\textwidth, trim=0 0 0 3em,clip]{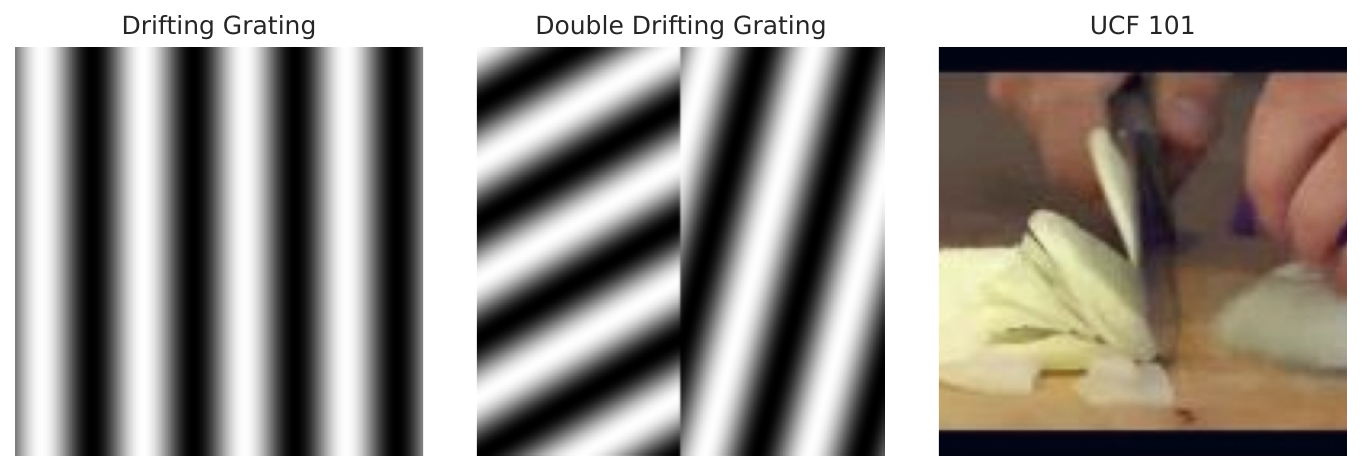}
\caption{\textbf{Linear probing across increasing target complexity.}
We probe embeddings from each block using labels that range from (a) low-level stimulus parameters to (b) cross-field relations to (c) high-level action categories.
Bottom: example frames from each dataset.}
\label{fig:feat_complexity}
\end{figure}

\subsection{Feature complexity analysis}
\label{sec:analysis:feat_complexity}
To probe what is linearly accessible at each depth, we decode targets that require progressively broader spatiotemporal integration using the same linear readout procedure on frozen embeddings.
We operationalize this as feature complexity: from local stimulus attributes, to cross-field relations, to naturalistic action semantics.
We use synthetic drifting gratings to obtain controlled, factorized targets (e.g., orientation, spatial/temporal frequency, contrast) without confounds from natural-video content.

\emph{Probe and protocol.}
All targets are categorical and decoded with multinomial logistic regression on frozen clip embeddings.
For synthetic grating datasets (unseen during pretraining), we report stratified cross-validated accuracy. For UCF101 we train on the official train split and evaluate on the test split.
Synthetic clips match the UCF101 clip format to minimize distribution shifts due to dataset. The probe setup matches Sec.~\ref{sec:analysis:downstream}. We use the following Target sets:
\textbf{Level 1: local attributes.}
Single full-field drifting gratings labeled by stimulus parameters (spatial \& temporal frequency, orientation, and contrast) (\figurename~\ref{fig:feat_complexity}a).\\
\textbf{Level 2: relational attributes.}
Two independently parameterized gratings in left/right hemifields, labeled by hemifield-specific parameters and relational targets (e.g., match/mismatch; signed differences) (\figurename~\ref{fig:feat_complexity}b).\\
\textbf{Level 3: downstream semantics.}
UCF101 action categories, which require integrating distributed spatiotemporal evidence beyond local parameters (\figurename~\ref{fig:feat_complexity}c).

Under a fixed probe, differences across blocks and regimes indicate how explicitly (and linearly) each level is organized and available in the embedding.

\subsection{Feature similarity across network depth}
\begin{figure}
\includegraphics[width=\textwidth, trim=0.8em 0 0.8em 0,clip]{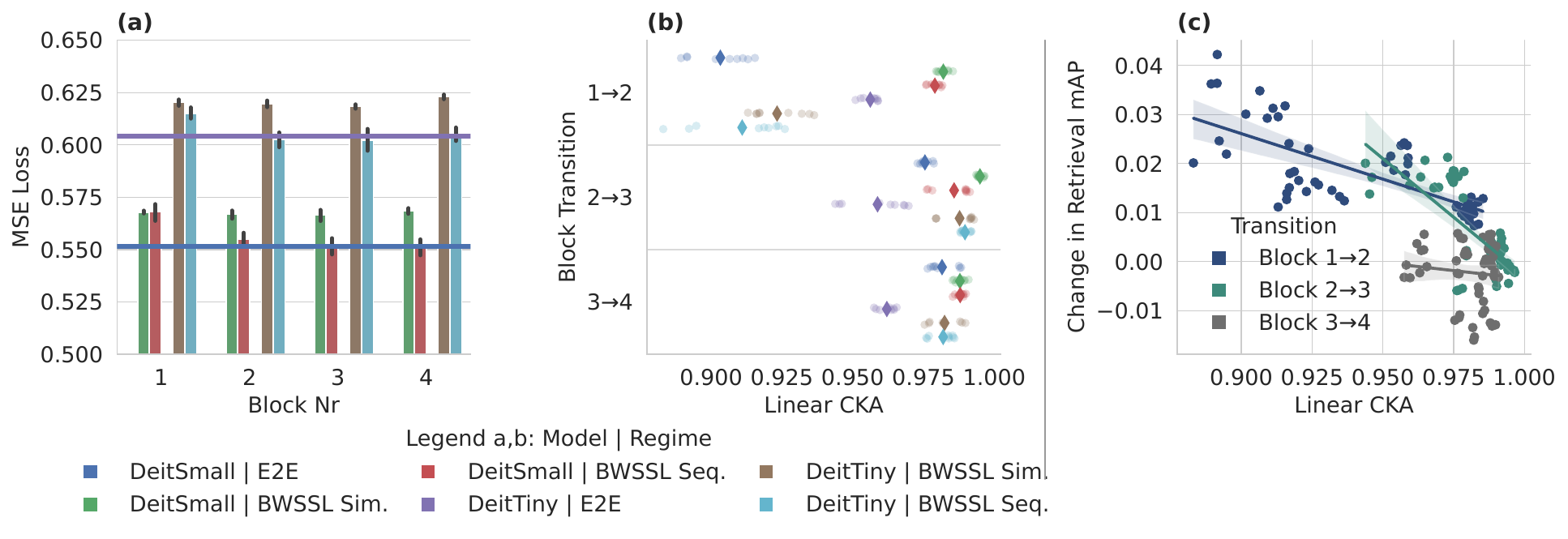}
\caption{\textbf{Reconstruction loss and inter-block similarity.}
(a) Masked reconstruction MSE after each block (E2E at the final block).
(b) CKA between successive blocks. Lower values indicate larger representational updates.
(c) Relationship between inter-block CKA and the corresponding change in retrieval mAP.}
\label{fig:cka}
\end{figure}

To quantify representational change with depth, we compute linear centered kernel alignment (CKA \cite{kornblithSimilarityNeuralNetwork2019a}) between successive blocks (\figurename~\ref{fig:cka}b).
We interpret high CKA between successive blocks as evidence of geometry preservation (consistent with feature reuse/refinement), whereas lower CKA indicates a larger representational update (consistent with re-encoding).

For each block $k$, we run the encoder on the UCF101 evaluation set (Sec.~\ref{sec:setup:data}) without masking and form $z_k\in\mathbb{R}^{D}$ by mean-pooling patch tokens.
Stacking the $N$ ($\approx 1100$) clip embeddings yields $X_k\in\mathbb{R}^{N\times D}$.
We compute
\[
\mathrm{CKA}(X_k, X_{k+1})
=\frac{\|X_k^\top X_{k+1}\|_F^2}{\|X_k^\top X_k\|_F\,\|X_{k+1}^\top X_{k+1}\|_F},
\]
after column-centering (subtracting the per-feature mean across clips). We also relate CKA to the change in retrieval mAP across blocks (Sec.~\ref{sec:analysis:downstream}) (\figurename~\ref{fig:cka}c).

\subsection{Patch-level detail retention and robustness}
\label{sec:analysis:patch_level}
We test whether task-relevant information remains localized or becomes globally mixed by measuring robustness to extreme spatial occlusion and the homogenization of patch-token representations.
Prior work suggests that patch tokens can become more similar with depth for ViT \cite{raghuVisionTransformersSee2021}.
We evaluate (i) accuracy degradation when most patches are removed and (ii) within-clip token similarity.

\paragraph{Occlusion sensitivity.}
We repeat our probing from Sec.~\ref{sec:analysis:feat_complexity} on embeddings from clips where all but a small set of patches are zeroed in pixel space prior to tokenization.
For single-grating clips we retain exactly one visible patch and for double-grating clips we retain two patches (one per hemifield) to keep left/right targets well-defined. Let $\mathrm{acc}_{\text{full}}$ be accuracy on full clips and $\mathrm{acc}_{\text{occl}}$ accuracy under occlusion. We report the relative drop:
\(
\mathrm{OccDrop} \;=\; \frac{\mathrm{acc}_{\text{full}}-\mathrm{acc}_{\text{occl}}}{\mathrm{acc}_{\text{full}}},
\)
averaged over patch locations and runs.
We omit UCF101 results since this occlusion drives all models to low accuracy and high run variability, rendering cross-model comparisons uninformative.

\paragraph{Token homogenization with cosine patch similarity (CPS).}
We measure within-clip homogenization at block $k$ via the mean pairwise cosine similarity of distinct patch tokens in $h_k$:
\(
\mathrm{CPS} \;=\; \mathbb{E}_{b}\big[ \mathrm{mean}_{i \neq j}( \cos(e_{b,i}, e_{b,j}) ) \big],
\)
computed after LayerNorm (without affine parameters) and $\ell_2$ normalization.
Higher CPS indicates more homogeneous (less patch-specific) token representations.
For efficiency, CPS is computed on a subsample of evaluation clips.

\begin{figure}
\includegraphics[width=\textwidth]{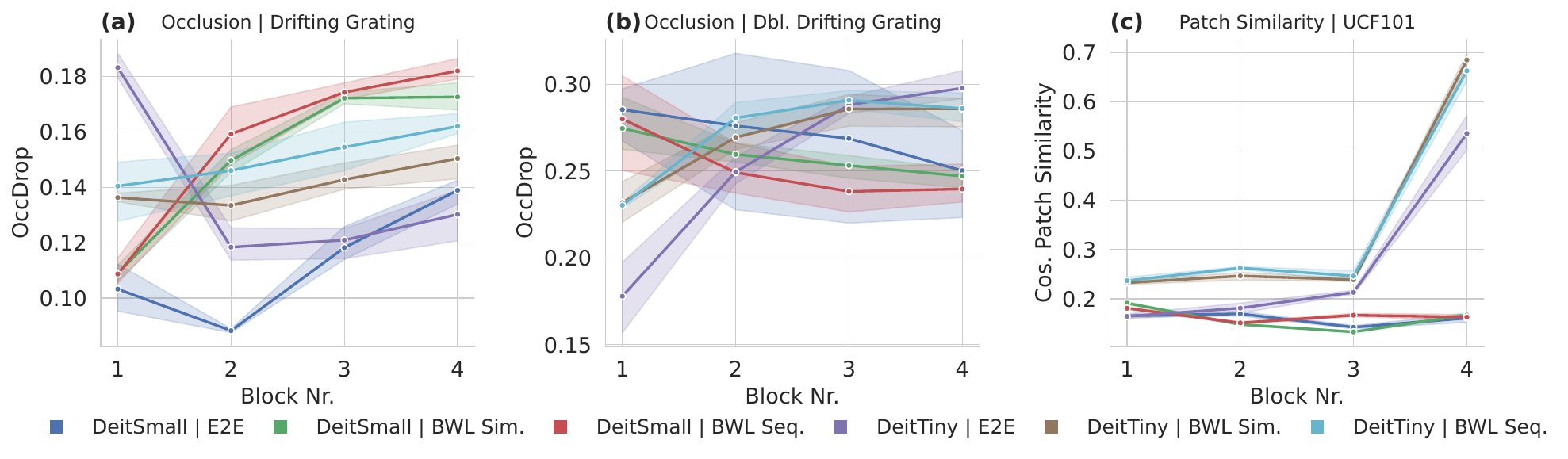}
\caption{\textbf{Patch-level robustness and token mixing across blocks.}
We quantify (a,b) robustness of linear decoding under extreme spatial occlusion (OccDrop in Sec.~\ref{sec:analysis:patch_level}) and (c) patch-token homogenization via cosine patch similarity (CPS).}
\label{fig:patch_level}
\end{figure}

\section{Results}

Our analysis of loss dynamics and representational characteristics across blocks revealed a number of distinct observations.

\paragraph{BWSSL VideoMAE on ViT backbones converges and approaches E2E representation quality.}
Across all settings, our BWSSL setup converged reliably.
MSE losses in the final-layer reconstruction were on par or slightly better in sequential BWSSL compared to E2E training. Simultaneous BWSSL typically yielded slightly higher MSE losses (\figurename~\ref{fig:key_metrics}c).
On downstream proxies, BWSSL remained close to E2E in both linear probing and retrieval mAP (\figurename~\ref{fig:key_metrics}a,b), with only small residual gaps, showing overall comparable linear accessibility of task-relevant information and the semantic structure of the embedding space.
For DeiT-Tiny, the gap was $\leq 0.02$ in linear-probe accuracy and $\leq 0.015$ in mAP, while for DeiT-Small, $\leq 0.04$ and $\leq 0.02$, respectively.
These consistent small gaps are in line with prior BWSSL results \cite{siddiqui2024blockwise,xiongLoCoLocalContrastive2020}.

\paragraph{Sequential and simultaneous training yield similar representations despite different loss dynamics.}
Simultaneous BWSSL reconstruction loss often plateaus after early blocks (\figurename~\ref{fig:cka}a), whereas sequential BWSSL shows more consistent improvement across depth. Since blocks are trained sequentially and then frozen, later blocks typically achieve better scores. 
Despite these differences, representation quality is nearly identical: across model sizes, block splits, and evaluation settings, sequential and simultaneous BWSSL usually differ by $<0.01$ in linear-probe Acc. and mAP (\figurename~\ref{fig:key_metrics}a,b), with neither variant consistently outperforming the other.
This similarity holds for depth-resolved evaluations (differences are typically within run-to-run variability): feature-complexity probes (\figurename~\ref{fig:feat_complexity}), representational similarity of blocks (\figurename~\ref{fig:cka}), and patch-level analyses (\figurename~\ref{fig:patch_level}).

\paragraph{Finer block granularity shows no notable effect on reconstruction loss.}
For both investigated model sizes, increasing the block granularity from four to six blocks did not significantly affect the final layer reconstruction loss, although the scores were overall slightly lower for DeiT-Small (\figurename~\ref{fig:key_metrics}c).
Representation quality of embeddings in the last module showed only small but consistent reductions with 6 modules (\figurename~\ref{fig:key_metrics}a,b): 
Retrieval mAP typically decreased by less than 0.01, and linear-probe accuracy by less then 0.02. Often, differences overlap within run-to-run variability. (\figurename~\ref{fig:key_metrics}a,b) 

\paragraph{BWSSL training promotes earlier linear access to higher-complexity information.}
Independent of model size, BWSSL decodes better than E2E at matched depth for relational and action targets (\figurename~\ref{fig:feat_complexity}b,c), suggesting that BWSSL makes mid- and high-level features linearly accessible in earlier layers.
At the same time, E2E training shows a stronger improvement with deeper modules and higher linear probe accuracy for low label complexity ~\ref{fig:feat_complexity}a). 
We thus see a shift in where higher-complexity targets are linearly accessible. BWSSL emphasizes accessibility in early modules, whereas E2E training gains more from later modules.

\paragraph{Low-complexity targets remain linearly decodable under BWSSL.}
Prior work on supervised blockwise learning reported losses of task-relevant information in early blocks with negative downstream effects \cite{wangInfoProLocallySupervised2025,sakamoto2024endtoend}.
In our self-supervised setting, E2E indeed yields higher low-level decoding than BWSSL at matched depth (\figurename~\ref{fig:feat_complexity}a), but the disadvantage is modest:
it is largest in the first block (about $0.02$ in accuracy) and does not systematically increase with depth, indicating only a small overall reduction in low-level linear accessibility.
The depth-wise trend depends on model size. For DeiT-Small, low-level decoding is nearly constant across blocks for both regimes, with an approximately stable gap. For DeiT-Tiny, decoding decreases with depth for all regimes, and the E2E-BWSSL gap narrows in later blocks as performance drops. Overall, we see no systematic amplification of low-level linear decodability decrease with depth under BWSSL.

\paragraph{BWSSL shows diminishing late-block gains alongside increasing inter-block similarity.}
While early blocks improve embedding quality under BWSSL, later blocks offer little and sometimes negative additional gains.
In particular, the block~3$\to$4 transition yields near-zero (and occasionally negative) marginal improvements in linear-probe performance (\figurename~\ref{fig:feat_complexity}) and MSE Loss (\figurename~\ref{fig:cka}a).
This is unlikely to be simple task saturation. Early BWSSL embeddings already approach the final BWSSL scores, yet still lag behind E2E indicating clear headroom that later BWSSL blocks fail to recover. This is reinforced by representational similarity across depth (\figurename~\ref{fig:cka}b). For DeiT-Small, BWSSL produces consistently high similarity between consecutive blocks (CKA $>0.98$) across transitions, whereas E2E shows a markedly lower first transition (block~1$\rightarrow$2: $\approx 0.9$), with similarity increasing only at greater depth.
For DeiT-Tiny, BWSSL yields lower CKA than E2E for the first transition. However, for later Blocks BWSSL CKA exceeds 0.98 again while E2E increases only slightly.
Overall, the depth at which BWSSL stops improving downstream performance coincides with a flattening of representational change across blocks. This link becomes even more evident in kNN retrieval (\figurename~\ref{fig:cka}c):
large $\Delta$mAP arises mainly in early transitions where consecutive representations are less similar, whereas later transitions (notably block~3$\rightarrow$4) concentrate at very high CKA with $\Delta$mAP near zero or negative. Taken together, these results indicate that once blockwise training reaches a high-similarity zone, subsequent modules largely preserve upstream structure instead of driving substantial representational reorganization, which is consistent with the diminishing returns observed in transfer metrics.

\paragraph{BWSSL can shift token mixing earlier, especially in smaller models.}
Patch-level diagnostics reveal structure not visible from pooled embeddings.
In DeiT-Tiny, BWSSL achieves higher CPS already in block~1 than E2E (\figurename~\ref{fig:patch_level}c), suggesting earlier within-clip token alignment. For BWSSL CPS changes little across intermediate blocks before rising again in the final block. For E2E it increases slightly with each block with a similar rise to BWSSL in the final block.
In DeiT-Small, CPS varies comparatively weakly with depth and differs little across regimes (\figurename~\ref{fig:patch_level}c), indicating that CPS is comparatively insensitive to the training rule at this model scale.
Overall, BWSSL can place the network in a distinct token-mixing configuration at the same depth, most clearly in the smaller model.

\paragraph{BWSSL can redistribute information across patch tokens.}
Occlusion sensitivity reveals task-dependent shifts in patch influence under BWSSL.
On drifting gratings, both BWSSL variants show larger relative performance drops than E2E from block~2 onward in both model sizes (\figurename~\ref{fig:patch_level}a), suggesting reduced influence of any single patch
On double drifting gratings, the pattern is less consistent and shows a lot of run variability. The depth-wise trend seems to be primarily based on model size rather than regime. However, there are clear differences between BWSSL and E2E for DeitTiny in the first two blocks (\figurename~\ref{fig:patch_level}b).
Together with CPS, these results indicate that BWSSL reshapes patch-level structure in ways that pooled metrics can obscure, affecting both locality and token mixing.

\paragraph{BWSSL tracks E2E more closely in smaller models.}
Across settings, the BWSSL-E2E gap increases with model size.
While DeiT-Tiny and DeiT-Small mostly show similar qualitative depth-wise trends, BWSSL is closer to E2E for DeiT-Tiny on accuracy and mAP. For DeiT-Small, both BWSSL variants remain more noticeably below the E2E baseline (\figurename~\ref{fig:key_metrics}a,b).
Patch-level diagnostics mirror this relation (\figurename~\ref{fig:patch_level}): DeiT-Tiny shows stronger early token alignment under BWSSL (higher block~1 CPS), whereas DeiT-Small exhibits weaker regime differences.

\section{Discussion}


\paragraph{Why can BWSSL remain near-competitive without long-range error propagation?}

BWSSL approaching E2E on our transfer proxies is notable but consistent with prior work showing that strong representations can emerge under BWSSL \cite{siddiqui2024blockwise}, so we do not attribute competitiveness to a single factor in our setup. Masked reconstruction provides dense per-block supervision: each block must output an embedding from which its decoder can reconstruct, discouraging interfaces that drop reconstruction-relevant cues. Transformer residual streams may further stabilize this depth-wise interface, making later refinement easier.
The remaining gap to E2E is plausibly due to missing global coordination: E2E can align early representations with the final embedding geometry, whereas BWL can converge to locally sufficient but not globally optimal interfaces.

\paragraph{How does BWSSL reshape the low-level $\leftrightarrow$ high-level accessibility profile across depth?}

BWSSL shifts high-level decodability earlier, likely because intermediate reconstruction losses reward early integration of non-local structure. Low-level decodability is only slightly reduced, consistent with redistribution or possibly slightly lower detail retention, rather than a strict trade-off. Mechanistically, this might be driven by supervision placement rather than gradient isolation per se.

\paragraph{What determines when block granularity helps?}

Increasing granularity introduces more boundaries but smaller per-block updates. This can make handoffs smoother (less chance of a large, irreversible jump in one module), while limiting how much new structure each block can build. In our setup, this trade-off appears to depend on model size.
A likely moderator is decoder strength: Since each block has its own decoder, the local objective can be met either by improving the embedding or by relying on decoder capacity. With decoder depth held fixed, increasing the number of modules makes decoders effectively stronger relative to each (shallower) block, favoring representations that require more relative decoder strength to reconstruct without reliably becoming more transfer-friendly.

\paragraph{Why do later modules sometimes add little beyond earlier representations?}

High consecutive similarity together with near-zero (or negative) marginal gains suggests that late BWL modules often operate in a ``reuse-and-refine'' mode rather than ``re-encode''. Two non-exclusive mechanisms are consistent with this pattern.

First, interface limitation: if an early module converges to an embedding that is sufficient for its local decoder but does not expose cues needed for downstream improvements, later modules can only transform what is already represented and may be unable to recover missing information. This implies a one-way constraint: locally adequate early interfaces can cap later gains.

Second, local-objective stabilization: even when relevant cues are present, a module may find it locally optimal to preserve incoming geometry because any exploratory change is immediately penalized by its own decoder. This also predicts schedule-dependent loss curves: in sequential training, later decoders can reduce reconstruction loss mainly by learning to decode an already-usable upstream embedding, without requiring large representational changes. In simultaneous training, the decoder’s input embedding keeps moving during co-optimization, which can blunt late loss improvements even when final transfer proxies are similar.

\paragraph{Why does model size interact with BWL in our setup?}

BWL is closer to E2E in DeiT-Tiny than in DeiT-Small on our proxy metrics, with corresponding differences in token-level diagnostics. A likely driver is a capacity-ratio change: the decoder is fixed, while the encoder size depends on the model, so the local reconstruction bottleneck is relatively much tighter in DeiT-Small. This alters what boundary embeddings must encode (and thus what is passed to the next block). 
Finally, the performance gap of BWSSL in regard to model size is likely affected by regime effects (dataset \& task complexity in relation to model size).

\paragraph{Implications and limitations.}
BWSSL with VideoMAE is compute- and memory-heavy because each block has its own decoder. We study it for learning dynamics rather than efficiency.
Methodologically, given the relatively modest pretraining dataset, the model-size interaction we report may be regime-dependent and should not be over-interpreted as a general scaling trend. Finally, BWSSL confounds gradient isolation with supervision placement by design. Our study does not aim to differentiate between gradient locality and deep supervision.

\section{Conclusion and Future Work}
We evaluated gradient-isolated blockwise self-supervised learning (BWSSL) for VideoMAE-style video ViTs. BWSSL converged and nearly matched end-to-end (E2E) training on linear-probe and retrieval proxies, but shifted depth dynamics. Higher-level information became linearly accessible earlier, whereas later blocks showed minimal representational change and correspondingly diminishing gains in performance. Our patch-level analyses revealed that BWSSL can lead to earlier token mixing and shifts in patch influence. These findings suggest that BWSSL is a viable but regime-sensitive alternative to E2E training for video ViTs, with performance and internal organization depending on objective–decoder design, partition granularity, and capacity.
Future work should identify interventions that prevent late-block saturation (e.g., decoder sharing, alternative local objectives) while reducing the compute and memory overhead.

\subsubsection{Acknowledgements}
The project received funding from the Helmholtz Association's Initiative and Networking Fund through the Helmholtz International BigBrain Analytics and Learning Laboratory (HIBALL) under the Helmholtz International Lab grant agreement InterLabs-0015, the Helmholtz Association portfolio theme "Supercomputing and Modeling for the Human Brain", and by the European Union's Horizon Europe Programme, grant agreement 101147319 (EBRAINS 2.0 Project). We used a large language model to optimize manuscript wording and facilitate code prototyping. All outputs were verified by the authors, who take full responsibility for the work.

%
%
%
\bibliography{main}

\end{document}